\documentclass[10pt,twocolumn,letterpaper]{article}

\usepackage{iccv}
\usepackage{times}
\usepackage{epsfig}
\usepackage{graphicx}
\usepackage[fleqn]{amsmath}
\usepackage{amssymb}
\usepackage{enumitem}
\usepackage{algorithm}
\usepackage{algpseudocode}
\usepackage{flushend}

\usepackage[pagebackref=true,breaklinks=true,letterpaper=true,colorlinks,bookmarks=false]{hyperref}

\iccvfinalcopy 

\usepackage{color}

\def\iccvPaperID{202} 

\ificcvfinal\fi
\begin{document}

\title{Visual Semantic Planning using Deep Successor Representations}

\author{Yuke Zhu$^{*}$$^3$\quad\quad Daniel Gordon\thanks{indicates equal contribution.}$^*$$^4$\quad\quad Eric Kolve$^1$\quad\quad Dieter Fox$^4$\\  Li Fei-Fei$^3$\quad\quad Abhinav Gupta$^{1,2}$\quad\quad Roozbeh Mottaghi$^1$\quad\quad Ali Farhadi$^{1,4}$\\
{\normalsize $^1$Allen Institute for Artificial Intelligence\quad$^2$Carnegie Mellon University}\\
{\normalsize $^3$Stanford University\quad}
{\normalsize $^4$University of Washington}\\
}

\maketitle
\thispagestyle{empty}

\begin{abstract}
A crucial capability of real-world intelligent agents is their ability to \textbf{plan} a sequence of actions to achieve their goals in the visual world. In this work, we address the problem of \textbf{visual semantic planning}: the task of predicting a sequence of actions from visual observations that transform a dynamic environment from an initial state to a goal state. Doing so entails knowledge about objects and their affordances, as well as actions and their preconditions and effects. We propose learning these through interacting with a visual and dynamic environment. Our proposed solution involves bootstrapping reinforcement learning with imitation learning. To ensure cross task generalization, we develop a deep predictive model based on successor representations. Our experimental results show near optimal results across a wide range of tasks in the challenging THOR environment. 


\end{abstract}

\section{Introduction}
 
Humans demonstrate levels of visual understanding that go well beyond current formulations of mainstream vision tasks (e.g. object detection, scene recognition, image segmentation). A key element to visual intelligence is the ability to interact with the environment and \textit{plan} a sequence of actions to achieve specific goals; This, in fact, is central to the survival of agents in dynamic environments~\cite{anderson03,noe2002}.

\begin{figure}[t!]
\begin{center}
\includegraphics[width=1.0\linewidth]{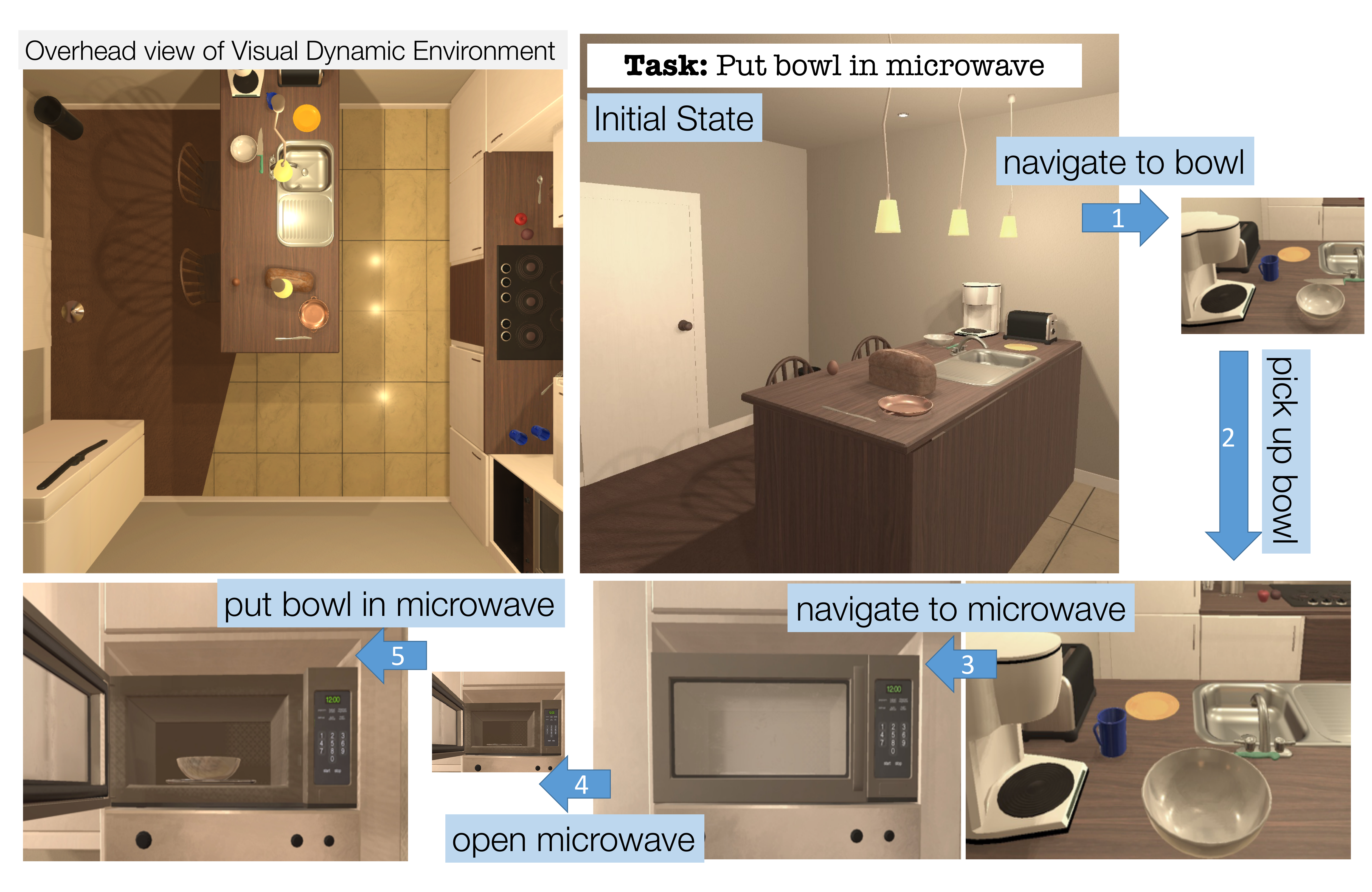}
\caption{Given a task and an initial configuration of a scene, our agent learns to interact with the scene and predict a sequence of actions to achieve the goal based on visual inputs.}\vspace{-6mm}
\label{fig:pull_figure}
\vspace{-3mm}
\end{center}
\end{figure}

\textit{Visual semantic planning}, the task of interacting with a visual world and predicting a sequence of actions that achieves a desired goal, involves addressing several challenging problems. For example, imagine the simple  task of \texttt{putting the bowl in the microwave} in the visual dynamic environment depicted in Figure~\ref{fig:pull_figure}. A successful plan involves first finding the bowl, navigating to it, then grabbing it, followed by finding and navigating to the microwave, opening the microwave, and finally putting the bowl in the microwave.

The \textbf{first} challenge in visual planning is that performing each of the above actions in a  visual dynamic environment requires deep visual understanding of that environment, including the set of possible actions, their preconditions and effects, and object affordances. For example, to \textit{open a microwave} an agent needs to know that it should be in front of the microwave, and it should be aware of the state of the microwave and not try to open an already opened microwave. Long explorations that are required for some tasks imposes the \textbf{second} challenge.
%
The variability of visual observations and possible actions makes na\"{i}ve exploration intractable. To \textit{find a cup}, the agent might need to search several cabinets one by one.
The \textbf{third} challenge is emitting a sequence of actions such that the agent ends in the goal state and the effects of the preceding actions meet the preconditions of the proceeding ones. Finally, a satisfactory solution to visual planning should enable cross task transfer; previous knowledge about one task should make it easier to learn the next one. This is the \textbf{fourth} challenge.

In this paper, we address \textit{visual semantic planning} as a policy learning problem. We mainly focus on high-level actions and do not take into account the low-level details of motor control and motion planning. \textbf{Visual Semantic Planning} (VSP) is the task of predicting a sequence of semantic actions that moves an agent from a random initial state in a visual dynamic environment to a given goal state.

To address the first challenge, one needs to find a way to represent the required knowledge of objects, actions, and the visual environment.  One possible way is to learn these from still images or videos~\cite{fathi13,tran08,Wang16}. But we argue that learning high-level knowledge about actions and their preconditions and effects requires an active and prolonged interaction with the environment. In this paper, we take an interaction-centric approach where we learn this knowledge through interacting with the \textit{visual dynamic environment}. Learning by interaction on real robots has limited scalability due to the complexity and cost of robotics systems~\cite{pinto2016curious,pinto2016supersizing,sutton1998reinforcement}. A common treatment is to use simulation as \emph{mental rehearsal} before real-world deployment~\cite{bellemare2013ALE,kempka2016vizdoom,lerer2016UETorch,wymann2015torcs,zhu2016target}. For this purpose, we use the THOR framework~\cite{zhu2016target}, extending it to enable interactions with objects, where an action is specified as its pre- and post-conditions in a formal language.


To address the second and third challenges, we cast VSP as a policy learning problem, typically tackled by reinforcement learning~\cite{dosovitskiy17,gu17,kim04,lillicrap2015continuous,mnih2015human,silver2016mastering}. To deal with the large action space and delayed rewards, we use imitation learning to bootstrap reinforcement learning and to guide exploration.  To address the fourth challenge of cross task generalization~\cite{lake2016building}, we develop a deep predictive model based on successor representations~\cite{dayan1993improving,kulkarni2016deep} that decouple environment dynamics and task rewards, such that knowledge from trained tasks can be transferred to new tasks with theoretical guarantees~\cite{barreto2016successor}.

In summary, we address the problem of visual semantic planning and propose an interaction-centric solution. Our proposed model obtains near optimal results across a spectrum of tasks in the challenging THOR environment. Our results also show that our deep successor representation offers crucial transferability properties. Finally, our qualitative results show that our learned representation can encode visual knowledge of objects, actions, and environments.

\section{Related Work}

\noindent\textbf{Task planning.} Task-level planning \cite{Dornhege09,fikes1971strips,kaelbling2011hierarchical,srivastava14,srivastava13} addresses the problem of finding a high-level plan for performing a task. These methods typically work with high-level formal languages and low-dimensional state spaces. In contrast, visual semantic planning is particularly challenging due to the high dimensionality and partial observability of visual input. In addition, our method facilitates generalization across tasks, while previous methods are typically designed for a specific environment and task. 
 
 \vspace{1mm}
\noindent\textbf{Perception and interaction.} Our work integrates perception and interaction, where an agent actively interfaces with the environment to learn policies that map pixels to actions. The synergy between perception and interaction has drawn an increasing interest in the vision and robotics community. Recent work has enabled faster learning and produced more robust visual representations~\cite{agrawal2016learning,malmir2015deep,pinto2016curious} through interaction. 
Some early successes have been shown in physical understanding~\cite{denil2016learning,lerer2016UETorch,li16tofall,mottaghi16} by interacting with an environment. 

 



\vspace{1mm}
\noindent
\textbf{Deep reinforcement learning.}  Recent work in reinforcement learning has started to exploit the power of deep neural networks. Deep RL methods have shown success in several domains such as video games~\cite{mnih2015human}, board games~\cite{silver2016mastering}, and continuous control~\cite{lillicrap2015continuous}. Model-free RL methods (e.g., \cite{lillicrap2015continuous,mnih2016asynchronous,mnih2015human}) aim at learning to behave solely from actions and their environment feedback; while model-based RL approaches (e.g., \cite{deisenroth11,schmidhuber1990line,tamar16}) also estimate a environment model. Successor representation (SR), proposed by Dayan~\cite{dayan1993improving}, can be considered as a hybrid approach of model-based and model-free RL. Barreto~\etal~\cite{barreto2016successor} derived a bound on value functions of an optimal policy when transferring policies using successor representations. Kulkarni~\etal~\cite{kulkarni2016deep} proposed a method to learn deep successor features. In this work, we propose a new SR architecture with significantly reduced parameters, especially in large action spaces, to facilitate model convergence. We propose to first train the model with imitation learning and fine-tune with RL. It enables us to perform more realistic tasks and offers significant benefits for transfer learning to new tasks.

\vspace{1mm}
\noindent
\textbf{Learning from demonstrations.}
Expert demonstrations offer a source of supervision in tasks which must usually be learned with copious random exploration. 
A line of work interleaves policy execution and learning from expert demonstration that has achieved good practical results~\cite{searn,dagger}.
Recent works have employed a series of new techniques for imitation learning, such as generative adversarial networks~\cite{ho2016generative,li2017infogail}, Monte Carlo tree search~\cite{guo2014deep} and guided policy search~\cite{levine2016end}, which learn end-to-end policies from pixels to actions.

\vspace{1mm}
\noindent
\textbf{Synthetic data for visual tasks.}  Computer games and simulated platforms have been used for training perceptual tasks, such as semantic segmentation~\cite{handa2016understanding}, pedestrian detection~\cite{marin2010learning}, pose estimation~\cite{papon2015semantic}, and urban driving~\cite{chen2015deepdriving,richter2016playing,ros2016synthia,shafaei2016play}. In robotics, there is a long history of using simulated environments for learning and testing before real-world deployment~\cite{kober2013reinforcement}. Several interactive platforms have been proposed for learning control with visual inputs~\cite{bellemare2013ALE,kempka2016vizdoom,lerer2016UETorch,wymann2015torcs,zhu2016target}. Among these, THOR~\cite{zhu2016target} provides high-quality realistic indoor scenes. Our work extends THOR with a new set of actions and the integration of a planner.




\begin{figure}
\includegraphics[width=1.0\linewidth]{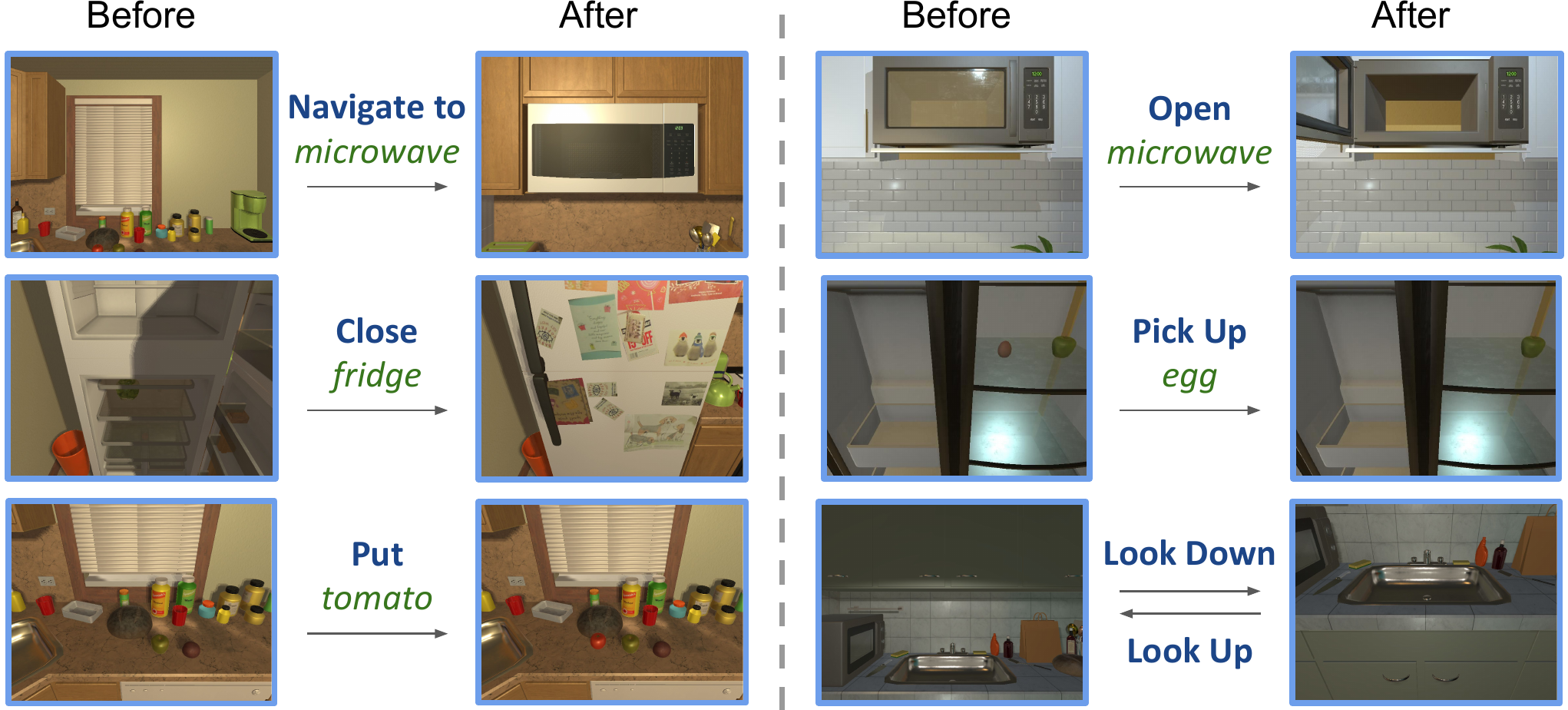}
\caption{Example images that demonstrate the state changes before and after an object interaction from each of the six action types in our framework. Each action changes the visual state and certain actions may enable further interactions such as opening the fridge before taking an object from it.}
\vspace{-3mm}
\label{fig:action_space}
\end{figure}

\section{Interactive Framework}
To enable interactions with objects and with the environment, we extend the THOR framework~\cite{zhu2016target}, which has been used for learning visual navigation tasks. Our extension includes new object states, and a discrete description of the scene in a planning language~\cite{fikes1971strips}.



\subsection{Scenes}
In this work, we focus on \emph{kitchen} scenes, as they allow for a variety of tasks with objects from many categories. Our extended THOR framework consists of 10 individual kitchen scenes. Each scene contains an average of 53 distinct objects with which the agent can interact. The scenes are developed using the Unity 3D game engine. 


\subsection{Objects and Actions}
\label{sec:action_space}
We categorize the objects by their affordances~\cite{gibson2014ecological}, i.e., the plausible set of actions that can be performed. For the tasks of interest, we focus on two types of objects: 1) \textit{items} that are small objects (mug, apple, etc.) which can be picked up, held, and moved by the agent to various locations in the scene, and 2) \textit{receptacles} that are large objects (table, sink, etc.) which are stationary and can hold a fixed capacity of \textit{items}. A subset of \textit{receptacles}, such as fridges and cabinets, are \textit{containers}. These \textit{containers} have doors that can be opened and closed. The agent can only put an \textit{item} in a \textit{container} when it is open. We assume that the agent can hold at most one \textit{item} at any point. We further define the following actions to interact with the objects:
\vspace{1mm}
\begin{enumerate}[noitemsep,nolistsep]
    \item \textsf{Navigate} \{\textit{receptacle}\}: moving from the current location of the agent to a location near the \textit{receptacle};
    \item \textsf{Open} $\{\textit{container}\}$: opening the door of a \textit{container} in front of an agent;
    \item \textsf{Close} $\{\textit{container}\}$: closing the door of a \textit{container} in front of an agent;
    \item \textsf{Pick\,\,Up} $\{\textit{item}\}$: picking up an \textit{item} in field of view;
    \item \textsf{Put} $\{\textit{receptacle}\}$: putting a held item in the \textit{receptacle};
    \item \textsf{Look\,\,Up} and \textsf{Look\,\,Down}: tilting the agent's gaze 30 degrees up or down.
\end{enumerate}
\vspace{1mm}
In summary, we have six action types, each taking a corresponding type of action arguments. The combination of actions and arguments results in a large action set of 80 per scene on average. Fig.~\ref{fig:action_space} illustrates example scenes and the six types of actions in our framework. Our definition of action space makes two important abstractions to make learning tractable: 1) it abstracts away from navigation, which can be tackled by a subroutine using existing methods such as \cite{zhu2016target}; and 2) it allows the model to learn with semantic actions, abstracting away from continuous motions, e.g., the physical movement of a robot arm to grasp an object. A common treatment for this abstraction is to ``fill in the gaps'' between semantic actions with callouts to a continuous motion planner~\cite{kaelbling2011hierarchical,srivastava14}. It is evident that not all actions can be performed in every situation. For example, the agent cannot pick up an \textit{item} when it is out of sight, or put a tomato into fridge when the fridge door is closed. To address these requirements, we specify the pre-conditions and effects of actions. Next we introduce an approach to declaring them as logical rules in a planning language. These rules are only encoded in the environment but not exposed to the agent. Hence, the agent must learn them through interaction.

\subsection{Planning Language}
\label{sec:planning}
The problem of generating a sequence of actions that leads to the goal state has been formally studied in the field of automated planning~\cite{ghallab2004automated}. Planning languages offer a standard way of expressing an automated planning problem instance, which can be solved by an off-the-shelf planner. We use STRIPS~\cite{fikes1971strips} as the planning language to describe our visual planning problem.

In STRIPS, a planning problem is composed of a description of an initial state, a specification of the goal state(s), and a set of actions. In visual planning, the initial state corresponds to the initial configuration of the scene. The specification of the goal state is a boolean function that returns true on states where the task is completed. Each action is defined by its precondition (conditions that must be satisfied before the action is performed) and postcondition (changes caused by the action). The STRIPS formulation enables us to define the rules of the scene, such as object affordances and causality of actions.


\begin{figure*}[t!]
\begin{center}
\includegraphics[width=.78\linewidth]{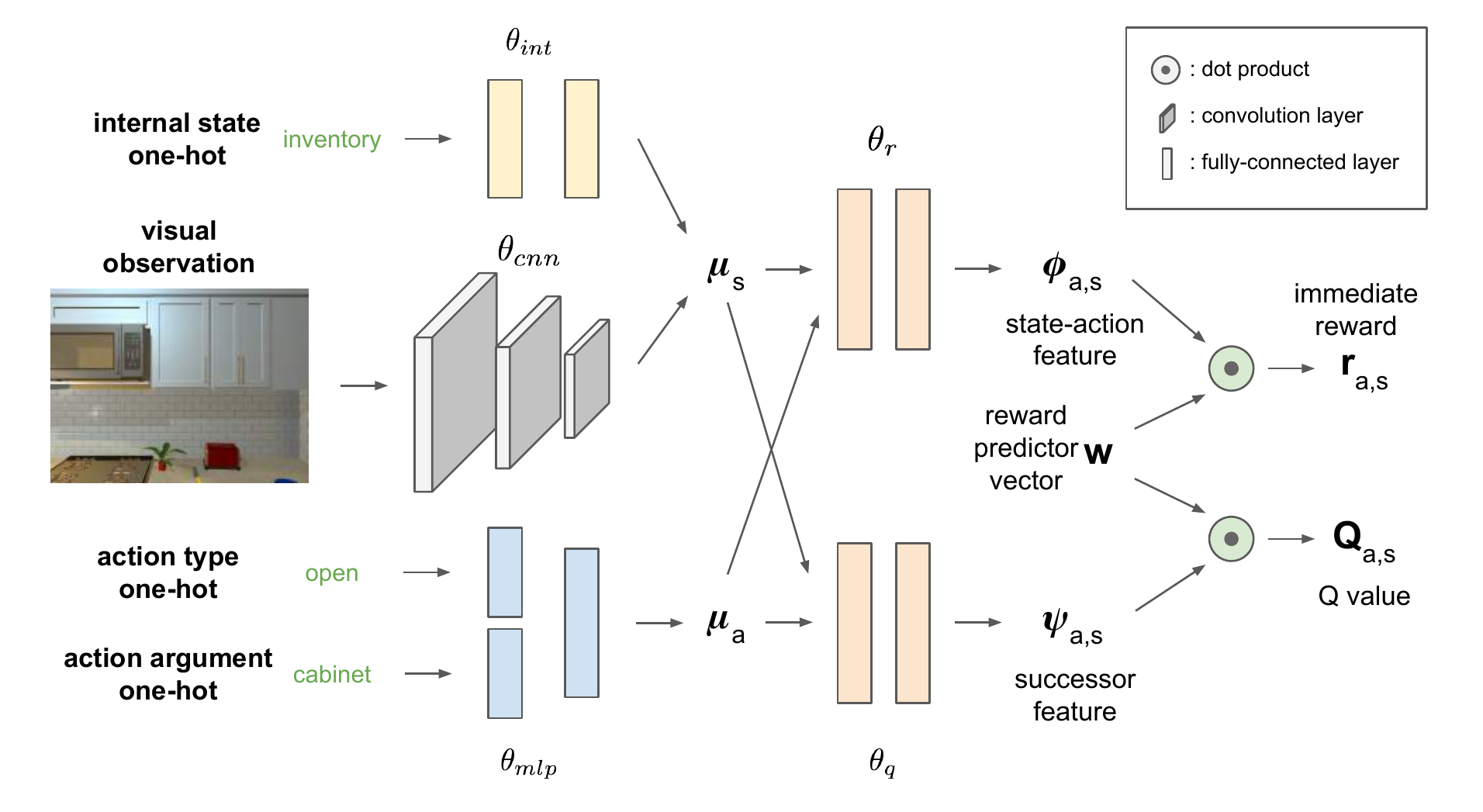}
\caption{An overview of the network architecture of our successor representation (SR) model. Our network takes in the current state as well as a specific action and predicts an immediate reward $r_{a,s}$ as well as a discounted future reward $Q_{a,s}$, performing this evaluation for each action. The learned policy $\pi$ takes the argmax over all $Q$ values as its chosen action.}
\vspace{-5mm}
\label{fig:architecture}
\end{center}
\end{figure*}

\section{Our Approach}
\label{sec:approach}
We first outline the basics of policy learning in Sec.~\ref{sec:background}. Next we formulate the \emph{visual semantic planning} problem as a policy learning problem and describe our model based on successor representation. Later we propose two protocols of training this model using imitation learning (IL) and reinforcement learning (RL). To this end, we use IL to bootstrap our model and use RL to further improve its performance.

\subsection{Successor Representation}
\label{sec:background}
We formulate the agent's interactions with an environment as a \emph{Markov Decision Process} (MDP), which can be specified by a tuple $(\mathcal{S}, \mathcal{A}, p, r, \gamma)$. $\mathcal{S}$ and $\mathcal{A}$ are the sets of states and actions. For $s\in\mathcal{S}$ and $a\in\mathcal{A}$, $p(s'|s, a)$ defines the probability of transiting from the state $s$ to the next state $s'\in\mathcal{S}$ by taking action $a$. $r(s,a)$ is a real-value function that defines the expected immediate reward of taking action $a$ in state $s$. For a state-action trajectory, we define the future discounted \emph{return} $R=\sum_{i=0}^{\infty}\gamma^i r(s_i, a_i)$, where $\gamma\in[0,1]$ is called the discount factor, which trades off the importance of immediate rewards versus future rewards.

A \emph{policy} $\pi:\mathcal{S}\rightarrow\mathcal{A}$ defines a mapping from states to actions. The goal of policy learning is to find the optimal policy $\pi^*$ that maximizes the future discounted return $R$ starting from state $s_0$ and following the policy $\pi^{*}$. Instead of directly optimizing a parameterized policy, we take a value-based approach. We define a state-action value function $Q^{\pi}:\mathcal{S}\times\mathcal{A}\rightarrow \mathbb{R}$ under a policy $\pi$ as
\begin{equation}
Q^{\pi}(s,a) = \mathbb{E}^{\pi}[R|s_0=s, a_0=a],
\label{eq:q_value}
\end{equation}
i.e., the expected episode return starting from state $s$, taking action $a$, and following policy $\pi$. The Q value of the optimal policy $\pi^{*}$ obeys the Bellman equation~\cite{sutton1998reinforcement}:
\begin{equation}
Q^{\pi^{*}}(s, a) = \mathbb{E}^{\pi^{*}}[r(s,a) + \gamma\max_{a'}Q(s',a')]
\label{eq:bellman}
\end{equation}
In deep Q networks~\cite{mnih2015human}, Q functions are approximated by a neural network $Q(s,a|\theta)$, and can be trained by minimizing the $\ell_2$-distance between both sides of the Bellman equation in Eq.~\eqref{eq:bellman}. Once we learn $Q^{\pi^{*}}$, the optimal action at state $s$ can be selected by $a^{*}=\arg\max_a Q^{\pi^{*}}(s,a)$.

Successor representation (SR), proposed by Dayan~\cite{dayan1993improving}, uses a similar value-based formulation for policy learning. It differs from traditional Q learning by factoring the value function into a dot product of two components: a reward predictor vector $\mathbf{w}$ and a predictive successor feature $\psi(s,a)$. To derive the SR formulation, we start by factoring the immediate rewards such that
\begin{equation}
r(s,a) = \phi(s,a)^T\mathbf{w},
\label{eq:immediate_reward}
\end{equation}
where $\phi(s,a)$ is a \emph{state-action feature}. We expand Eq.~\eqref{eq:q_value} using this reward factorization:
\begin{align}
Q^{\pi}(s,a)= &\,~ \mathbb{E}^{\pi}[ \sum_{i=0}^{\infty}\gamma^i r(s_i,a_i)|s_0=s, a_0=a]\nonumber \\
 = &\,~ \mathbb{E}^{\pi}[ \sum_{i=0}^{\infty}\gamma^i \phi(s_i,a_i)^T\mathbf{w}|s_0=s, a_0=a]\nonumber \\
 = &\,~ \mathbb{E}^{\pi}[ \sum_{i=0}^{\infty}\gamma^i \phi(s_i,a_i)|s_0=s, a_0=a]^T\mathbf{w} \nonumber\\
 = &\,~ \psi^{\pi}(s,a)^T\mathbf{w} \label{eq:succ_reward}
\end{align}
We refer to $\psi(s,a)^{\pi} = \mathbb{E}^{\pi}[ \sum_{i=0}^{\infty}\gamma^i \phi_{s,a}|s_0=s, a_0=a]$ as the \emph{successor features} of the pair $(s,a)$ under policy $\pi$. 

Intuitively, the successor feature $\psi^{\pi}(s,a)$ summarizes the environment dynamics under a policy $\pi$ in a state-action feature space, which can be interpreted as the expected future ``feature occupancy''. The reward predictor vector $\mathbf{w}$ induces the structure of the reward functions, which can be considered as an embedding of a task. Such decompositions have been shown to offer several advantages, such as being adaptive to changes in distal rewards and apt to option discovery~\cite{kulkarni2016deep}.
A theoretical result derived by Barreto~\etal implies a bound on performance guarantee when the agent transfers a policy from a task $t$ to a similar task $t'$, where task similarity is determined by the $\ell_2$-distance of the corresponding $\mathbf{w}$ vectors between these two tasks $t$ and $t'$~\cite{barreto2016successor}. Successor representation thus provides a generic framework for policy transfer in reinforcement learning.

\subsection{Our Model}

We formulate the problem of \emph{visual semantic planning} as a policy learning problem.  Formally, we denote a task by a Boolean function $t:\mathcal{S}\rightarrow \{0,1\}$, where a state $s$ completes the task $t$ iff $t(s)=1$. The goal is to find an optimal policy $\pi^*$, such that given an initial state $s_0$, $\pi^*$ generates a state-action trajectory $\mathcal{T}=\{(s_i, a_i)\,|\,i=0\dots T\}$ that maximizes the sum of immediate rewards $\sum_{i=0}^{T-1}r(s_i, a_i)$, where $t(s_{0\ldots T-1})=0$ and $t(s_T)=1$.

We parameterize such a policy using the successor representation (SR) model from the previous section. We develop a new neural network architecture to learn $\phi$, $\psi$ and $\mathbf{w}$. The network architecture is illustrated in Fig.~\ref{fig:architecture}. In THOR, the agent's observations come from a first-person RGB camera. We also pass the agent's internal state as input, expressed by one-hot encodings of the held object in its inventory. The action space is described in Sec.~\ref{sec:action_space}. We start by computing embedding vectors for the states and the actions. The image is passed through a 3-layer convolutional encoder, and the internal state through a 2-layer MLP, producing a state embedding $\mu_s=f(s;\theta_{cnn},\theta_{int})$. The action $a=[a_{type},a_{arg}]$ is encoded as one-hot vectors and passed through a 2-layer MLP encoder that produces an action embedding $\mu_a=g(a_{type},a_{arg};\theta_{mlp})$. We fuse the state and action embeddings and generate the state-action feature $\phi_{s,a}=h(\mu_s,\mu_a;\theta_r)$ and the successor feature $\psi_{s,a}=m(\mu_s,\mu_a;\theta_q)$ in two branches. The network predicts the immediate reward $r_{s,a}=\phi_{s,a}^T\mathbf{w}$ and the Q value under the current policy $Q_{s,a}=\psi_{s,a}^T\mathbf{w}$ using the decomposition from Eq.~\eqref{eq:immediate_reward} and \eqref{eq:succ_reward}.

\subsection{Imitation Learning}
\begin{figure}[t]
\begin{center}
\includegraphics[width=1.0\linewidth]{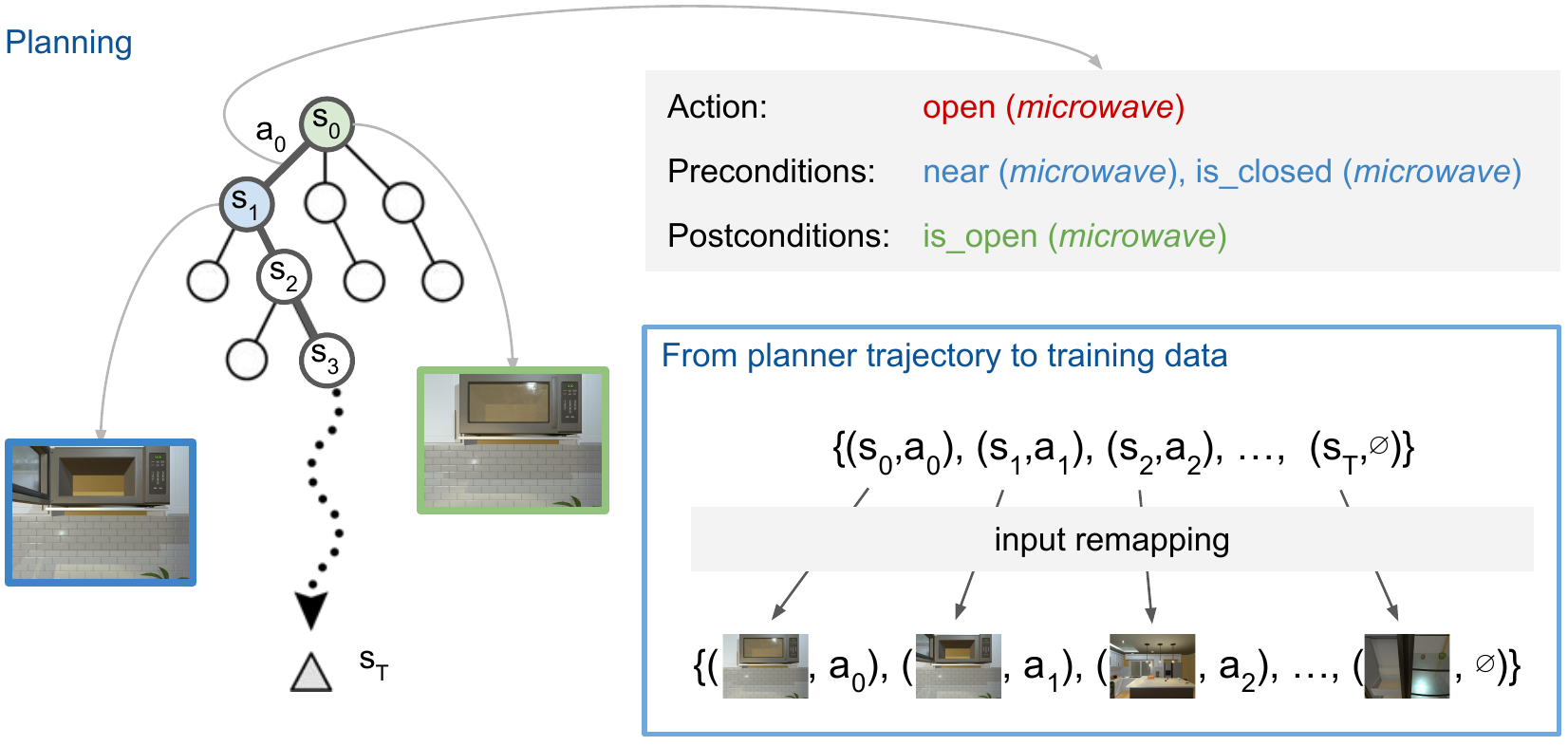}
\caption{We use a planner to generate a trajectory from an initial state-action pair $(s_0, a_0)$ to a goal state $s_T$. We describe each scene in a STRIPS-based planning language, where actions are specified by their pre- and post-conditions (see Sec.~\ref{sec:planning}). We perform input remapping, illustrated in the blue box, to obtain the image-action pairs from the trajectory as training data. After performing an action, we update the plan and repeat.}
\vspace{-4mm}
\label{fig:planning}
\end{center}
\end{figure}

Our SR-based policy can be learned in two fashions. First, it can be trained by imitation learning (IL) under the supervision of the trajectories of an optimal planner. Second, it can be learned by trial and error using reinforcement learning (RL). In practice, we find that the large action space in THOR makes RL from scratch intractable due to the challenge of exploration. The best model performance is produced by IL bootstrapping followed by RL fine-tuning. Given a task, we generate a state-action trajectory:
\begin{equation}
\mathcal{T} = \{(s_0,a_0),\{(s_1,a_1),\ldots,(s_{T-1},a_{T-1}), (s_T,\emptyset)\}
\label{eq:imitation_learning_reward}
\end{equation}
using the planner from the initial state-action pair  ($s_0, a_0$) to the goal state $s_T$ (no action is performed at terminal states). This trajectory is generated on a low-dimensional state representation in the STRIPS planner (Sec.~\ref{sec:planning}). Each low-dimensional state corresponds to an RGB image, i.e., the agent's visual observation. During training, we perform \emph{input remapping} to supervise the model with image-action pairs rather than feeding the low-dimensional planner states to the network. To fully explore the state space, we take planner actions as well as random actions off the optimal plan. After each action, we recompute the trajectory. This process of generating training data from a planner is illustrated in Fig.~\ref{fig:planning}.
Each state-action pair is associated with a true immediate reward $\hat{r}_{s,a}$. We use the mean squared loss function to minimize the error of reward prediction:
\begin{equation}
\mathcal{L}_r=\frac{1}{T}\sum_{i=0}^{T-1}(\hat{r}_{s,a} - \phi_{s,a}^T\mathbf{w})^2.
\label{eq:loss}
\end{equation}
Following the REINFORCE rule~\cite{sutton1998reinforcement}, we use the discounted return along the trajectory $\mathcal{T}$ as an unbiased estimate of the true Q value: $\hat{Q}_{s,a}\approx\sum_{i=0}^{T-1} \gamma^i\hat{r}_{s,a}$. We use the mean squared loss to minimize the error of Q prediction:
\begin{equation}
\mathcal{L}_Q=(\hat{Q}_{s,a} - \psi_{s,a}^T\mathbf{w})^2
\end{equation}
The final loss on the planner trajectory $\mathcal{T}$ is the sum of the reward loss and the Q loss: $\mathcal{L}_{\mathcal{T}}=\mathcal{L}_r+\mathcal{L}_Q$. Using this loss signal, we train the whole SR network on a large collection of planner trajectories starting from random initial states.

\subsection{Reinforcement Learning}
When training our SR model using RL, we can still use the mean squared loss in Eq.~\eqref{eq:loss} to supervise the learning of reward prediction branch for $\phi$ and $\mathbf{w}$. However, in absence of expert trajectories, we would need an iterative way to learn the successor features $\psi$. Rewriting the Bellman equation in Eq.~\eqref{eq:bellman} with the SR factorization, we can obtain an equality on $\phi$ and $\psi$:
\begin{equation}
\psi^{\pi^{*}}(s, a) = \mathbb{E}^{\pi^{*}}[\phi(s,a) + \gamma \psi(s',a')]
\label{eq;succ_update}
\end{equation}
where $a' = \arg\max_a \psi(s',a)^T\mathbf{w}$. Similar to DQN~\cite{mnih2015human}, we minimize the $\ell_2$-loss between both sides of Eq.~\eqref{eq;succ_update}:
\begin{equation}
L_{SR}= \mathbb{E}^{\pi}[(\phi_{s,a} + \gamma \psi_{s',a'} - \psi^{\pi}_{s, a})^2]
\label{eq;succ_update_loss}
\end{equation}
We use a similar procedure to Kulkarni~\etal~\cite{kulkarni2016deep} to train our SR model. The model alternates training between the reward branch and the SR branch. At each iteration, a mini-batch is randomly drawn from a replay buffer of past experiences~\cite{mnih2015human} to perform one SGD update.

\subsection{Transfer with Successor Features}
A major advantage of successor features is its ability to transfer across tasks by exploiting the structure shared by the tasks. Given a fixed state-action representation $\phi$, let $M_{\phi}$ be the set of all possible MDPs induced by $\phi$ and all instantiations of the reward prediction vectors $\mathbf{w}$. Assume that $\pi_i^*$ is the optimal policy  of the $i$-th task in the set $\{M_i\in M_{\phi}|i=1,\ldots n\}$. Let $M_{n+1}$ to be a new task. We denote $Q_{n+1}^{\pi_i^*}$ as the value function of executing the optimal policy of the task $M_i$ on the new task $M_{n+1}$, and $\tilde{Q}_{n+1}^{\pi_i^*}$ as an approximation of $Q_{n+1}^{\pi_i^*}$ by our SR model. Given a bound on the approximations such that
\begin{equation*}
|Q_{n+1}^{\pi_i^*}(s,a) - \tilde{Q}_{n+1}^{\pi_i^*}(s,a)| \leq \epsilon\quad \forall s\in\mathcal{S}, a\in\mathcal{A}, i=1,\ldots,n,
\end{equation*}
we define a policy $\pi'$ for the new task $M_{n+1}$ using $\tilde{Q}_{1,\dots,n}$, where $\pi'(s)=\arg\max_a \max_i \tilde{Q}_{n+1}^{\pi_i^*}(s,a)$. Theorem 2 in Barreto~\etal~\cite{barreto2016successor} implies a bound of the gap between value functions of the optimal policy $\pi_{n+1}^{*}$ and the policy $\pi'$:
\begin{equation*}
\small
Q_{n+1}^{\pi_{n+1}^*}(s,a) - Q_{n+1}^{\pi'}(s,a) \leq \frac{2\phi_{m}}{1-\gamma}(\min_i ||\mathbf{w}_i-\mathbf{w}_{n+1}|| + \epsilon),
\end{equation*}
where $\phi_m=\max_{s,a}||\phi(s,a)||$. This result serves the theoretical foundation of policy transfer in our SR model. In practice, when transferring to a new task while the scene dynamics remain the same, we freeze all model parameters except the single vector $\mathbf{w}$. This way, the policy of the new task can be learned with substantially higher sample efficiency than training a new network from scratch.

\subsection{Implementation Details}
We feed a history of the past four observations, converted to grayscale, to account for the agent's motions. 
We use a time cost of $-0.01$ to encourage shorter plans and a task completion reward of $10.0$.
We train our model with imitation learning for 500k iterations with a batch size of 32, and a learning rate of 1e-4. We also include the successor loss in Eq.~\eqref{eq;succ_update_loss} during imitation learning, which helps learn better successor features. We subsequently fine-tune the network with reinforcement learning with 10,000 episodes.

\section{Experiments}
We evaluate our model using the extended THOR framework on a variety of household tasks. We compare our method against standard reinforcement learning techniques as well as with non-successor based deep models. The tasks compare the different methods' abilities to learn across varying time horizons. We also demonstrate the SR network's ability to efficiently adapt to new tasks. Finally, we show that our model can learn a notion of object affordance by interacting with the scene.

\subsection{Quantitative Evaluation}
\label{sec:quantitative_eval}
We examine the effectiveness of our model and baseline methods on a set of tasks that require three levels of planning complexity in terms of optimal plan length.
\vspace{-2mm}
\paragraph{Experiment Setup}
We explore the two training protocols introduced in Sec.~\ref{sec:approach} to train our SR model:
\begin{enumerate}
\vspace{-1mm}
\item \textbf{RL}: we train the model solely based on trial and error, and learn the model parameters with RL update rules.
\vspace{-1mm}
\item \textbf{IL}: we use the planner to generate optimal trajectories starting from a large collection of random initial state-action pairs. We use the imitation learning methods to train the networks using supervised losses.
\end{enumerate}


\begin{table*}[]
\begin{center}
\resizebox{\textwidth}{!}{
\begin{tabular}{l|cc|cc|cc}
\hline
                                   & \multicolumn{2}{c|}{\textbf{Easy}}                      & \multicolumn{2}{c|}{\textbf{Medium}}                    & \multicolumn{2}{c}{\textbf{Hard}}                      \\ 
                                   & \textbf{Success Rate} & \textbf{Mean ($\sigma$) Episode Length} & \textbf{Success Rate} & \textbf{Mean ($\sigma$) Episode Length} & \textbf{Success Rate} & \textbf{Mean ($\sigma$) Episode Length} \\ \hline
                                   \hline
                                   
Random Action                      & 1.00         & 696.33 (744.71)                & 0.00         & -                              & 0.04         & 2827.08 (927.84)               \\
Random Valid Action                & 1.00         & 64.03 (68.04)                  & 0.02         & 3897.50 (548.50)               & 0.36         & 2194.83 (1401.72)              \\
A3C \cite{mnih2016asynchronous}    & 0.96         & 101.12 (151.04)                & 0.00         & -                              & 0.04         & 2674.29 (4370.40)              \\
CLS-MLP                            & 1.00         & \textbf{2.42} (0.70)           & 0.65         & 256.32 (700.78)                & 0.65         & 475.86 (806.42)                \\
CLS-LSTM                           & 1.00         & 2.86 (0.37)                    & \textbf{0.80}         & 314.05 (606.25)       & \textbf{0.66}& 136.94 (523.60)                \\
SR IL (ours)                       & 1.00         & 2.70 (1.06)                    & \textbf{0.80}         & 32.32 (29.22)         & 0.65         & \textbf{34.25} (63.81)                  \\
SR IL + RL (ours)                  & 1.00         & 2.57 (1.04)                    & \textbf{0.80}         & \textbf{26.56} (3.85) & -            & -                              \\ \hline
Optimal planner                    & 1.00         & 2.36 (1.04)                    & 1.00         & 12.10 (6.16)                   & 1.00         & 14.13 (9.09)                   \\ \hline
\end{tabular}
}
\end{center}
\vspace{-2mm}
\caption{Results of evaluating the model on the easy, medium, and hard tasks. For each task, we evaluate how many out of the 100 episodes were completed (success rate) and the mean and standard deviation for successful episode lengths. The numbers in parentheses show the standard deviations. We do not fine-tune our SR IL model for the hard task.}
\vspace{-2mm}
\label{table:quantitative}
\end{table*}

Empirically, we find that training with reinforcement learning from scratch cannot handle the large action space. Thus, we report the performance of our SR model trained with imitation learning (SR IL) as well as with additional reinforcement learning fine-tuning (SR IL\,+\,RL).

We compare our SR model with the state-of-the-art deep RL model, A3C~\cite{mnih2016asynchronous}, which is an advantage-based actor-critic method that allows the agent to learn from multiple copies of simulation while updating a single model in an asynchronous fashion. 
A3C establishes a strong baseline for reinforcement learning. We further use the same architecture to obtain two imitation learning (behavior cloning) baselines. 
We use the same A3C network structure to train a softmax classifier that predicts the planner actions given an input. The network predicts both the action types (e.g., \textsf{Put}) and the action arguments (e.g., \emph{apple}). We call this baseline CLS-MLP. We also investigate the role of memory in these models. To do this, we add an extra LSTM layer to the network before action outputs, called CLS-LSTM. We also include simple agents that take random actions and take random valid actions at each time step.

\vspace{-2mm}
\paragraph{Levels of task difficulty}
We evaluate all of the models with three levels of task difficulty based on the length of the optimal plans and the source of randomization:
\begin{enumerate}
\vspace{-1mm}
\item \textbf{Level 1 (Easy)}: \textsf{Navigate} to a \textit{container} and toggle its state. A sample task would be \texttt{go to the microwave and open it if it is closed, close it otherwise.} The initial location of the agent and all \emph{container} states are randomized. This task requires identifying object states and reasoning about action preconditions.

\vspace{-1mm}
\item \textbf{Level 2 (Medium)}: \textsf{Navigate} to multiple \textit{receptacles}, collect \textit{items}, and deposit them in a \textit{receptacle}. A sample task here is \texttt{pick up three mugs from three cabinets and put them in the sink.} Here we randomize the agent's initial location, while the item locations are fixed. This task requires a long trajectory of correct actions to complete the goal.

\vspace{-1mm}
\item \textbf{Level 3 (Hard)}: Search for an \textit{item} and put it in a \textit{receptacle}. An example task is \texttt{find the apple and put it in the fridge.} We randomize the agent's location as well as the location of all items. This task is especially difficult as it requires longer-term memory to account for partial observability, such as which cabinets have previously been checked.

\end{enumerate}
We evaluate all of the models on 10 easy tasks, 8 medium tasks, and 7 hard tasks, each across 100 episodes. Each episode terminates when a goal state is reached. We consider an episode fails if it does not reach any goal state within 5,000 actions. We report the episode success rate and mean episode length as the performance metrics. We exclude these failed episodes in the mean episode length metric. For the easy and medium tasks, we train the imitation learning models to mimic the optimal plans. However for the hard tasks, imitating the optimal plan is infeasible, as the location of the object is uncertain. In this case, the target object is likely to hide in a cabinet or a fridge which the agent cannot see. Therefore, we train the models to imitate a plan which searches for the object from all the \emph{receptacles} in a fixed order. For the same reason, we do not perform RL fine-tuning for the hard tasks.

\begin{figure}[t]
\begin{center}
\includegraphics[width=1.0\linewidth]{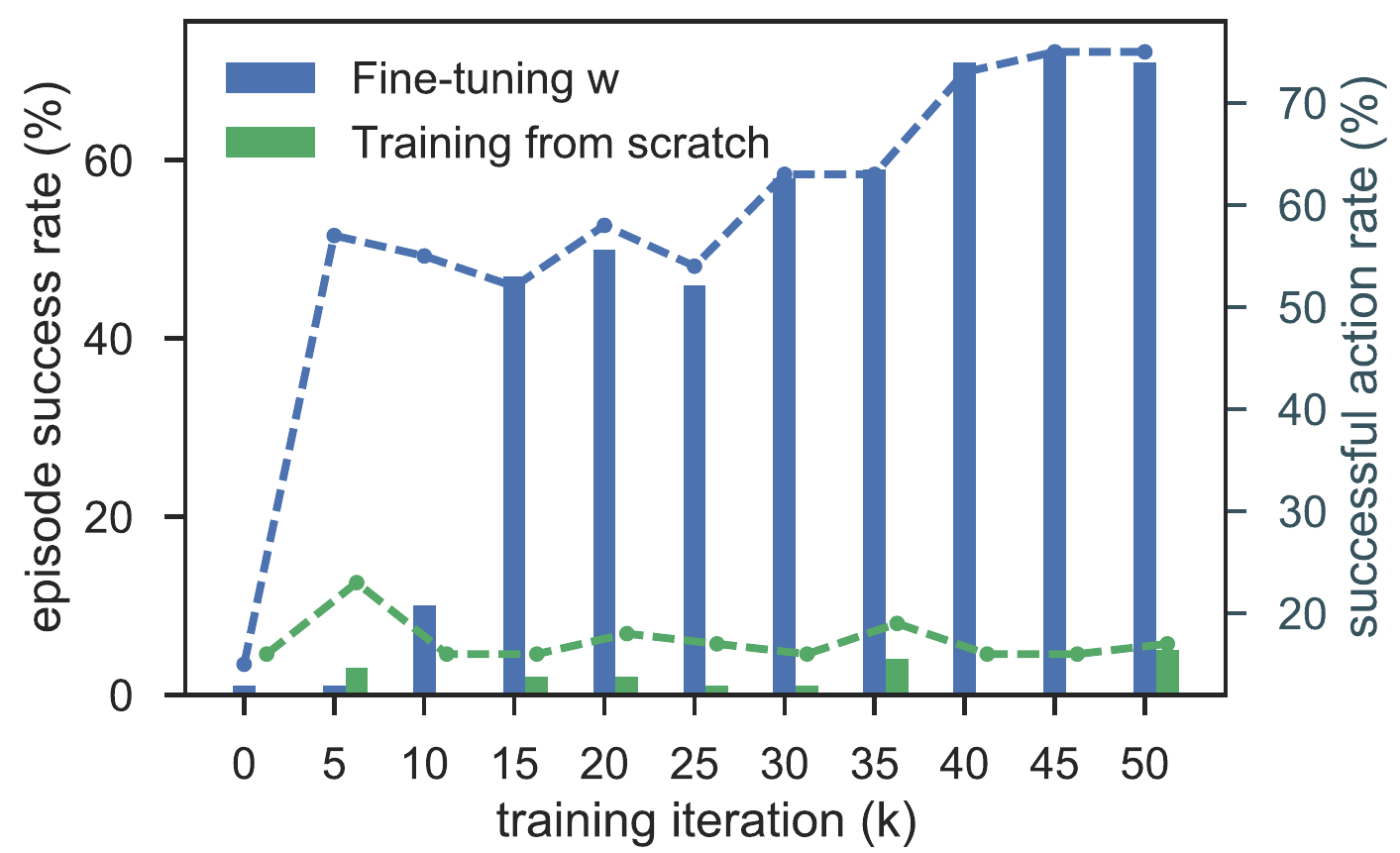}
\caption{We compare updating $\mathbf{w}$ with retraining the whole network for new hard tasks in the same scene. By using successor features, we can quickly learn an accurate policy for the new item. Bar charts correspond to the episode success rates, and line plots correspond to successful action rate.} 
\vspace{-7mm}
\label{fig:transfer}
\end{center}
\end{figure}

Table \ref{table:quantitative} summarizes the results of these experiments.
Pure RL-based methods struggle with the medium and hard tasks because the action space is so large that na\"{i}ve exploration rarely, if ever, succeeds. 
Comparing CLS-MLP and CLS-LSTM, adding memory to the agent helps improving success rate on medium tasks as well as completing tasks with shorter trajectories in hard tasks. Overall, the SR methods outperform the baselines across all three task difficulties. Fine-tuning the SR IL model with reinforcement learning further reduces the number of steps towards the goal. More qualitative results can be found in the video.\footnote{Link to supplementary video: \url{https://goo.gl/vXsbQP}}

\subsection{Task Transfer}
One major benefit of the successor representation decomposition is its ability to transfer to new tasks while only retraining the reward prediction vector $\mathbf{w}$, while freezing the successor features. We examine the sample efficiency of adapting a trained SR model on multiple novel tasks in the same scene. We examine policy transfer in the hard tasks, as the scene dynamics of the searching policy retains, even when the objects to be searched vary.
We evaluate the speed at which the SR model converges on a new task by fine-tuning the $\mathbf{w}$ vector versus training the model from scratch. We take a policy for searching a bowl in the scene and substituting four new items (lettuce, egg, container, and apple) in each new task. Fig.~\ref{fig:transfer} shows the episode success rates (bar chart) and the successful action rate (line plot). 
By fine-tuning $\mathbf{w}$, the model quickly adapts to new tasks, yielding both high episode success rate and  successful action rate. In contrast, the model trained from scratch takes substantially longer to converge. We also experiment with fine-tuning the entire model, and it suffers from similar slow convergence.


\begin{figure}[t]
\begin{center}
\includegraphics[width=.9\linewidth]{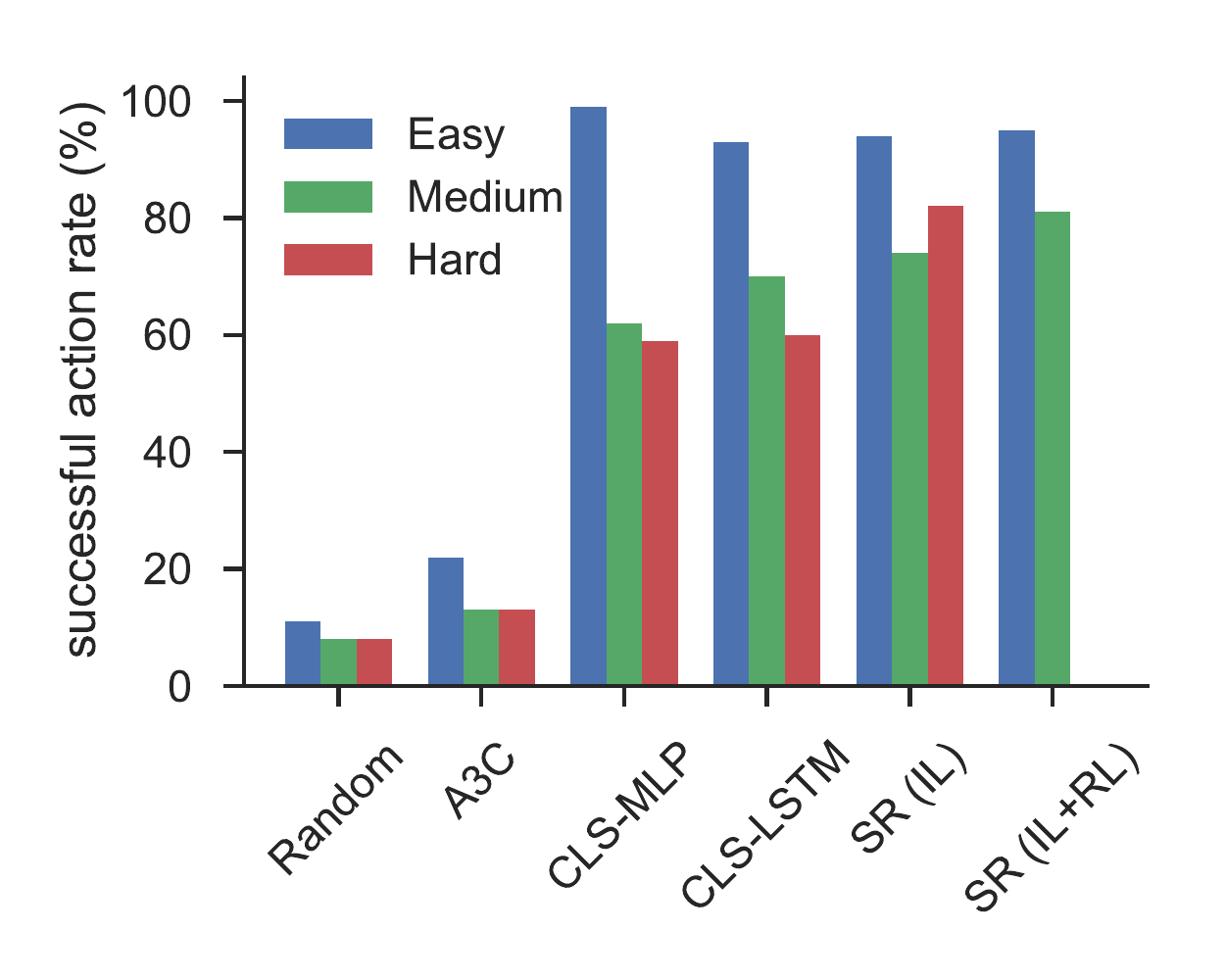}
\vspace{-4mm}
\caption{We compare the different models' likelihood of performing a successful action during execution. A3C suffers from the large action space due to na\"{i}ve exploration. Imitation learning models are capable of differentiating between successful and unsuccessful actions because the supervised loss discourages the selection of unsuccessful actions.} 
\vspace{-7mm}
\label{fig:affordance}
\end{center}
\end{figure}

\subsection{Learning Affordances}
An agent in an interactive environment needs to be able to reason about the causal effects of actions. We expect our SR model to learn the pre- and post-conditions of actions through interaction, such that it develops a notion of affordance~\cite{gibson2014ecological}, i.e., which actions can be performed under a circumstance. In the real world, such knowledge could help prevent damages to the agent and the environment caused by unexpected or invalid actions. 

We first evaluate each network's ability to \textit{implicitly} learn affordances when trained on the tasks in Sec.~\ref{sec:quantitative_eval}. In these tasks, we penalize unnecessary actions with a small time penalty, but we do not explicitly tell the network which actions succeed and which fail. Fig.~\ref{fig:affordance} illustrates that a standard reinforcement learning method cannot filter out unnecessary actions especially given delayed rewards. Imitation learning methods produce significantly fewer failed actions because they can directly evaluate whether each action gets them closer to the goal state.

\begin{figure}[t]
\begin{center}
\includegraphics[width=1.0\linewidth]{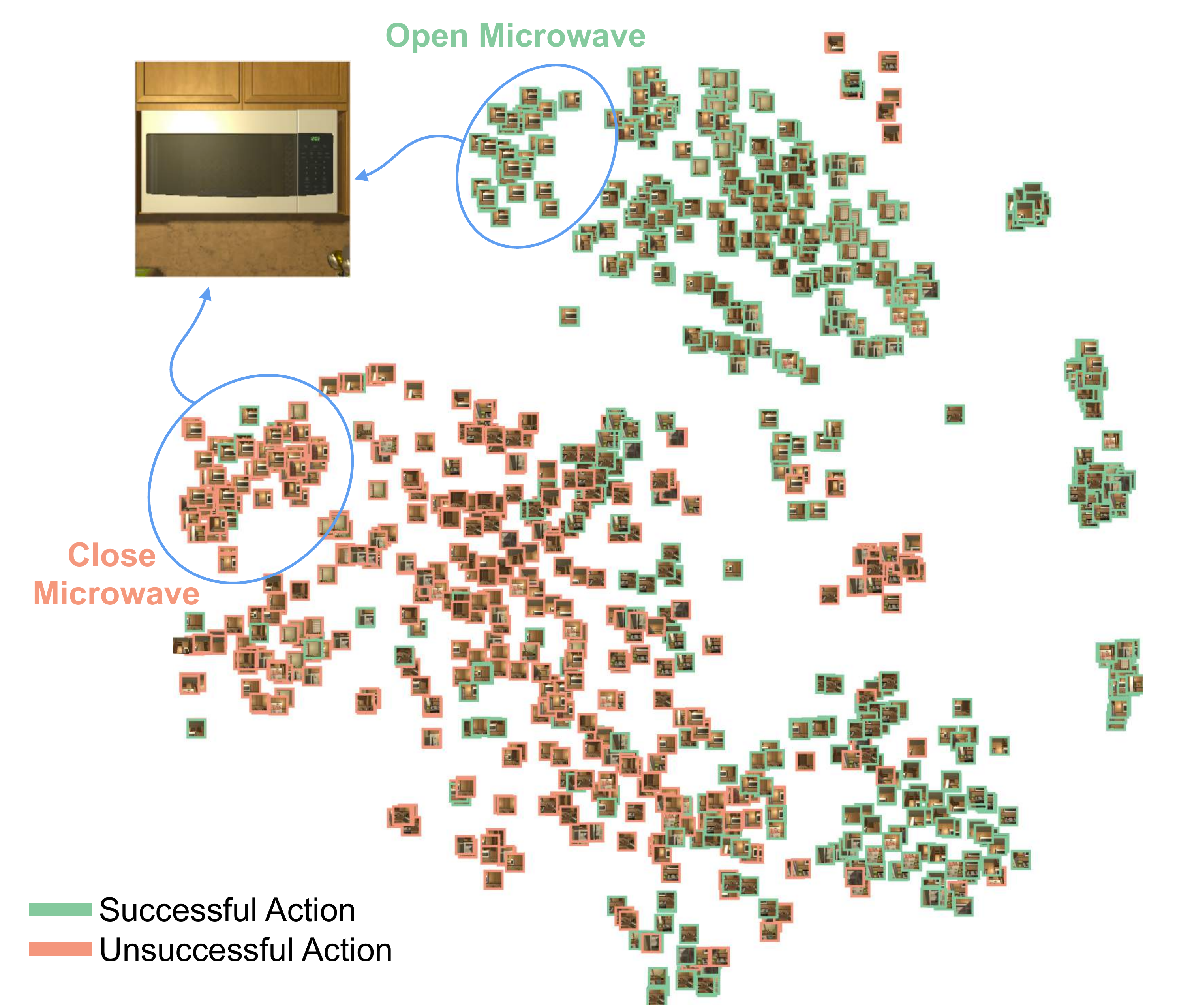}
\caption{Visualization of a t-SNE \cite{tsne} embedding of the state-action vector  $\phi_{s,a}$ for a random set of state-action pairs. Successful state-action pairs are shown in green, and unsuccessful pairs in orange. The two blue circles highlight portions of the embedding with very similar images but different actions. The network can differentiate successful pairs from unsuccessful ones.}
\vspace{-7mm}
\label{fig:tsne}
\end{center}
\end{figure}

We also analyze the successor network's capability of \textit{explicitly} learning affordances. We train our SR model with reinforcement learning, by executing a completely random policy in the scene. We define the immediate reward of issuing a successful action as $+1.0$ and an unsuccessful one as $-1.0$. The agent learns in 10,000 episodes. Fig.~\ref{fig:tsne} shows a t-SNE~\cite{tsne} visualization of the state-action features $\phi_{s,a}$. We see that the network learns to cluster successful state action pairs (shown in green) separate from unsuccessful pairs (orange). The network achieves an ROC-AUC of 0.91 on predicting immediate rewards over random state-action actions, indicating that the model can differentiate successful and unsuccessful actions by performing actions and learning from their outcomes. 
\section{Conclusions}
In this paper, we argue that visual semantic planning is an important next task in computer vision. Our proposed solution shows promising results in predicting a sequence of actions that change the current state of the visual world to a desired goal state. We have examined several different tasks with varying degrees of difficulty and show that our proposed model based on deep successor representations achieves near optimal results in the challenging THOR environment. We also show promising cross-task knowledge transfer results, a crucial component of any generalizable solution. Our qualitative results show that our learned successor features encode knowledge of object affordances, and action pre-conditions and post-effects. Our next steps involve exploring knowledge transfer from THOR to real-world environments as well as examining the possibilities of more complicated tasks with a richer set of actions.


\subsection*{Acknowledgements}
\noindent
This work is in part supported by ONR N00014-13-1-0720, ONR MURI N00014-16-1-2007, NSF IIS-1338054, NSF-1652052, NRI-1637479, NSF IIS-1652052, a Siemens grant, the Intel Science and Technology Center for Pervasive Computing (ISTC-PC), Allen Distinguished Investigator Award, and the Allen Institute for Artificial Intelligence.
{\small
\bibliographystyle{ieee}
\bibliography{egbib}
}

\appendix
\clearpage
\newpage
\section{Experiment Details}

\subsection{Experiment Setup}
We used the Adam optimizer from (Kingma and Ba) for learning our Successor Representation (SR) model with a learning rate of 1e-4 and a mini-batch size of 32. For the reinforcement learning experiments, we use the discounted factor $\gamma=0.99$ and a replay buffer size of 100,000. The exploration term $\epsilon$ is annealed from 1.0 to 0.1 during the training process. We run an $\epsilon$-greedy policy ($\epsilon=0.1$) during evaluation. We use soft target updates ($\tau=0.1$) after every episode. For the easy and medium tasks, we assign $+10.0$ immediate reward for task completion, $-5.0$ for invalid actions, and $-1.0$ for other actions (to encourage a shorter plan). For the hard task, we train our SR model to imitate a plan that searches all the \emph{receptacles} for an object in a fixed order of visitation based on the spatial locations of the \emph{receptacles}. We assign $+1.0$ immediate reward for task completion, and an episode terminates as failure if the agent does not follow the order of visitation in the plan. 

\subsection{Network Inputs}
The input to the SR model consists of three components: action (action type and argument), agent's observation (image), and agent's internal state. The action type is encoded by a 7-dimensional one-hot vector, indicating one of the seven action types (\textsf{Navigate}, \textsf{Open}, \textsf{Close}, \textsf{Pick Up}, \textsf{Put}, \textsf{Look Up}, and \textsf{Look Down}). The action argument is encoded by a one-hot vector that has the same dimension as the number of interactable objects plus one. The first dimension denotes null argument used for \textsf{Look Up} and \textsf{Look Down} actions, and the other dimensions correspond to the index of each object. RGB images from the agent's first-person camera are preprocessed to $84\times 84$ grayscale images. We stack four history frames to make an $84\times84\times4$ tensor as the image input to the convolutional networks. The agent's internal state is expressed by the agent's inventory, rotation, and viewpoint. The inventory is a one-hot vector that represents the index of the held \emph{item}, with an extra dimension for null. The rotation is a 4-dimensional one-hot vector that represents the rotation of the agent (90 degree turns). The viewpoint is a 3-dimensional one-hot vector that represents the tiling angle of the agent's camera ($-30^{\circ}$, $0^{\circ}$, and $30^{\circ}$).

\subsection{Network Architecture}
Here we describe the network architecture of our proposed SR model.
The convolutional image encoder $\theta_{cnn}$ takes an $84\times84\times4$ image as input. The three convolutional layers are 32 filters of size $8\times8$ with stride $4$, 64 filters of size $4\times4$ with stride $2$, 64 filters of size $3\times3$ with stride $1$. Finally a fully-connected layer maps the outputs from the convolutional encoder into a 512-d feature. The actions encoder $\theta_{mlp}$ and internal state encoder $\theta_{int}$ are both 2-layer MLPs with 512 hidden units. A concatenated vector of action, internal state, and image encodings is fed into two 2-layer MLPs $\theta_{r}$ and $\theta_{q}$ with 512 hidden units to produce the 512-d state-action feature $\phi_{s,a}$ and the successor feature $\psi_{s,a}$. We take the dot product of the 512-d reward predictor vector $\mathbf{w}$ and state-action features (successor features) to compute the immediate rewards (Q values). All the hidden layers use ReLU non-linearities. The final dot product layers of the immediate reward and the Q value produce raw values without any non-linearity.

\section{Algorithm Details}
We describe the reinforcement learning procedure of the SR model in Algorithm~\ref{algo:deep_sr}. This training method follows closely with previous work on deep Q-learning~\cite{mnih2015human} and deep SR model~\cite{kulkarni2016deep}. Similar to these two works, replay buffer and target network are used to stabilize training.

\begin{algorithm*}
  \caption{Reinforcement Learning for Successor Representation Model}
  \label{algo:deep_sr}
  \begin{algorithmic}[1]
  \Procedure{RL-Training}{}
  \State Initialize replay buffer $\mathcal{D}$ to size $N$
  \State Initialize an SR network $\mathcal{\theta}$ with random weights $\theta = [\theta_{int}, \theta_{cnn}, \theta_{mlp}, \theta_{r}, \theta_{q}, \mathbf{w}]$
  \State Make a clone of $\theta$ as the target network $\tilde{\theta}$
  \For{$i=1:$ \emph{\#episodes}}:
    \State Initialize an environment with random configuration
    \State Reset exploration term $\epsilon=1.0$
    \While {not terminal}
      \State Get agent's observation and internal state $s_t$ from the environment
      \State Compute $Q_{s_t,a}=f(s_t, a; \theta)$ for every action $a$ in action space
      \State With probability $\epsilon$ select a random action $a_t$; otherwise, select $a_t=\arg\max_aQ_{s_t,a}$
      \State Execute action $a_t$ to obtain the immediate reward $r_t$ and the next state $s_{t+1}$
      \State Store transition $(s_t, a_t, r_t, s_{t+1})$ in $\mathcal{D}$
      \State Sample a random mini-batch of transitions $(s_j, a_j, r_j, s_{j+1})$ from $\mathcal{D}$
      \State Compute $\tilde{r}_j$, $\phi_{s_j,a_j}$, and $\psi_{s_j,a_j}$ using $\theta$ for every transition $j$
      \State Compute gradients that minimize the mean squared error between $r_j$ and $\tilde{r}_j$
      \State Compute $\phi_{s_{j+1}, a}$, $\psi_{s_{j+1}, a}$, and $\tilde{Q}_{s_{j+1}, a}$ using $\tilde{\theta}$ for every transition $j$ and every action $a$
      \If {$s_{j+1}$ is a terminal state}:
        \State Compute gradients that minimize the mean squared error between  $\psi_{s_j, a_j}$ and $\phi_{s_j, a_j}$
      \Else:
        \State Compute gradients that minimize the mean squared error between $\psi_{s_j, a_j}$ and $\phi_{s_j, a_j} + \gamma \psi_{s_{j+1},a'}$
        \State where $a'=\arg\max_a \tilde{Q}_{s_{j+1}, a}$
      \EndIf
      \State Perform a gradient descend step to update $\theta$
    \EndWhile
    \State Anneal exploration term $\epsilon$
    \State Soft-update target network $\tilde{\theta}$ using $\theta$
  \EndFor
  \EndProcedure
  \end{algorithmic}
\end{algorithm*}

\section{Action Space}
The set of plausible actions in a scene is determined by the variety of objects in the scene. On average each scene has 53 objects (a subset of them are interactable) and the agent is able to perform 80 actions.  Here we provide an example scene to illustrate the interactable objects and the action space.

\vspace{4mm}
\noindent
\textbf{Scene \#9}:
16 \emph{items}, 23 \emph{receptacles} (at 11 unique locations), and 15 \emph{containers} (a subset of \emph{receptacles})

\begin{figure}[htbp]
\begin{center}
\includegraphics[width=1.0\linewidth]{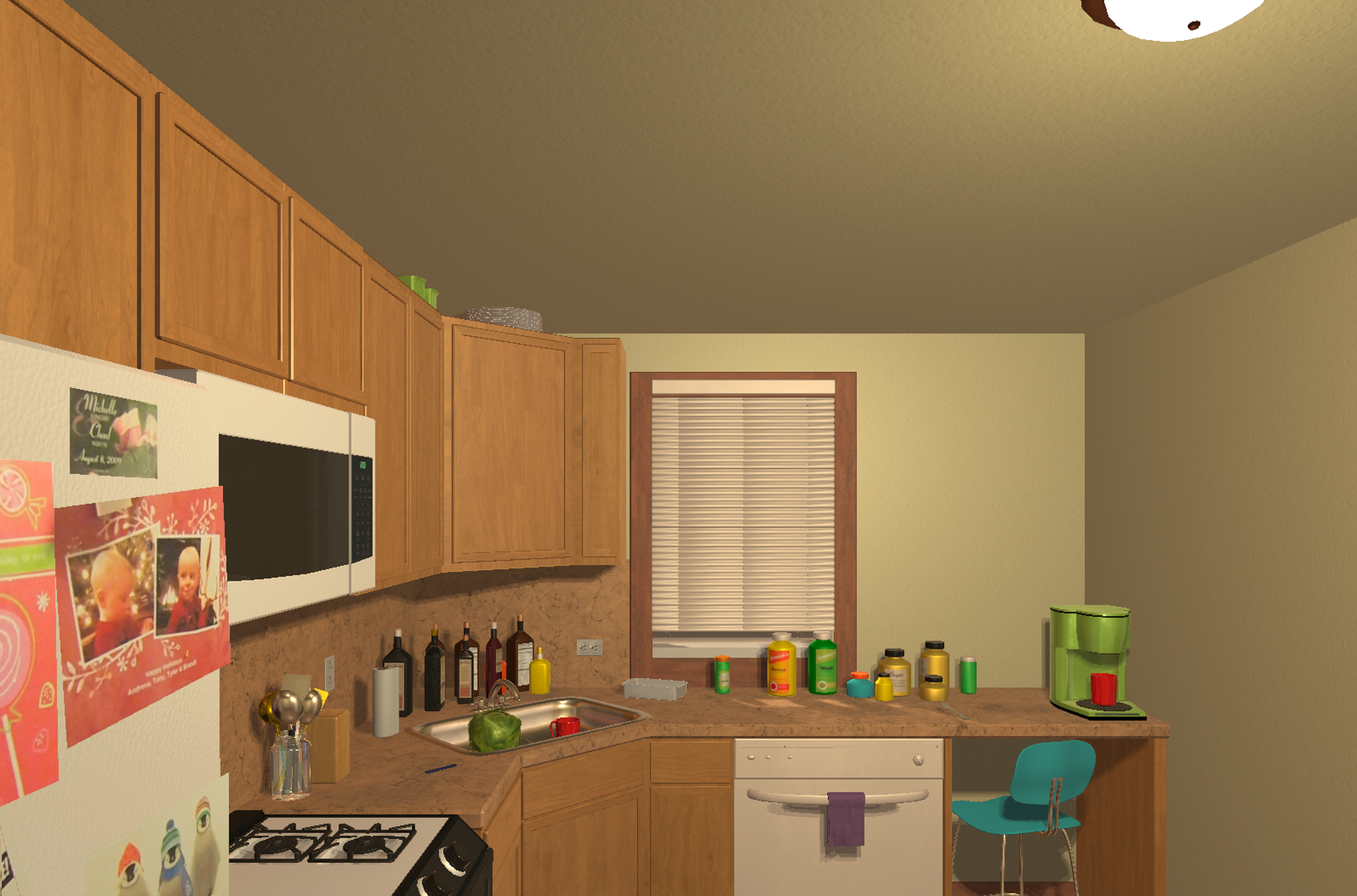}
\caption{Screenshot of Scene \#9}
\label{fig:scene_9}
\end{center}
\end{figure}
\vspace{-3mm}

\noindent
\textbf{items:} \emph{apple}, \emph{bowl}, \emph{bread}, \emph{butter knife}, \emph{glass bottle}, \emph{egg}, \emph{fork}, \emph{knife}, \emph{lettuce}, \emph{mug 1-3}, \emph{plate}, \emph{potato}, \emph{spoon}, \emph{tomato}

\vspace{0.5mm}
\noindent
\textbf{receptacles:} \emph{cabinet 1-13}, \emph{coffee machine}, \emph{fridge}, \emph{garbage can}, \emph{microwave}, \emph{sink}, \emph{stove burner 1-4}, \emph{table top}

\vspace{0.5mm}
\noindent
\textbf{containers:} \emph{cabinet 1-13}, \emph{fridge}, \emph{microwave}

\vspace{0.5mm}
\noindent
\textbf{actions:} 80 actions in total, including 11 \textsf{Navigation} actions, 15 \textsf{Open} actions, 15 \textsf{Close} actions, 14 \textsf{Pick Up} actions, 23 \textsf{Put} actions, \textsf{Look Up} and \textsf{Look Down}.

\vspace{2mm}

\noindent
We have fewer \textsf{Navigation} and \textsf{Pick Up} actions than the number of receptacles and items respectively, as we merge some adjacent receptacles to one location (navigation destination). We also merge picking up items from the same object category into one action. This reduces the size of the action space and speeds up learning. An important simplification that we made is to treat the \textsf{Navigation} actions as ``teleports'', which abstracts away from visual navigation of the agent. The actual visual navigation problem can be solved as an independent subroutine from previous work~\cite{zhu2016target}. 
As discussed in Sec.~\ref{sec:action_space}, not all actions in the set can be issued given a certain circumstance based on affordance. We use the PDDL language to check if the preconditions of an action are satisfied before the action is sent to THOR for execution.

\section{Tasks}
We list all the tasks that we have evaluated in the experiments in Table~\ref{table:task_list}. In summary, we evaluated tasks from three levels of difficulty, with 10 easy tasks, 8 medium tasks, and 7 hard tasks. 

\begin{table*}[htp]
\caption{List of Tasks from Three Levels of Difficulty}
\vspace{1mm}
\begin{center}
\begin{tabular}{clll}
\hline
\textbf{Scene} & \textbf{Easy} & \textbf{Medium} & \textbf{Hard}\\
\hline\hline
1 & open\,/\,close \emph{fridge} & put \emph{lettuce, tomato} and \emph{glass bottle} to the \emph{sink} & find \emph{bowl} and put in \emph{sink}\\
2 & open\,/\,close \emph{cabinet} & put \emph{apple, egg} and \emph{glass bottle} to the \emph{table top} & find \emph{plate} and put in \emph{cabinet}\\
3 & open\,/\,close \emph{microwave} & put \emph{glass bottle, lettuce} and \emph{apple} to  the \emph{table top} & find \emph{lettuce} and put in \emph{fridge}\\
4 & open\,/\,close \emph{cabinet} & put three \emph{mug}s to the \emph{fridge} & find \emph{glass bottle} and put in \emph{microwave}\\
5 & open\,/\,close \emph{fridge} & - & -\\
6 & open\,/\,close \emph{fridge} & - & -\\
7 & open\,/\,close \emph{cabinet} & put three \emph{mug}s to the \emph{table top} & -\\
8 & open\,/\,close \emph{fridge} & put \emph{potato, tomato} and \emph{apple} to the \emph{sink} & find \emph{lettuce} and put on \emph{table top}\\
9 & open\,/\,close \emph{microwave} & put three \emph{mug}s to the \emph{table top} & find \emph{glass bottle} and put in \emph{fridge}\\
10 & open\,/\,close \emph{cabinet} & put \emph{glass bottle, bread} and \emph{lettuce} to \emph{fridge} & find \emph{bowl} and put in \emph{sink}\\
\hline
\end{tabular}
\end{center}
\label{table:task_list}
\end{table*}%

\end{document}